\documentclass[10pt]{article}
\usepackage{colacl}
\usepackage{graphicx}
\title{Summarizing Encyclopedic Term Descriptions on the Web}
\author
{Atsushi Fujii and Tetsuya Ishikawa \\
Graduate School of Library, Information and Media Studies\\
University of Tsukuba \\
1-2 Kasuga, Tsukuba, 305-8550, Japan \\
\{fujii,ishikawa\}@slis.tsukuba.ac.jp}

\newenvironment{abstractquote}{\begin{list}{}{
\setlength{\rightmargin}{0.25in}\setlength{\leftmargin}{0.25in}}\item[]}{\end{list}}

\begin{document}
\maketitle

\begin{abstract}
\begin{abstractquote}
  We are developing an automatic method to compile an encyclopedic
  corpus from the Web. In our previous work, paragraph-style
  descriptions for a term are extracted from Web pages and organized
  based on domains. However, these descriptions are independent and do
  not comprise a condensed text as in hand-crafted encyclopedias. To
  resolve this problem, we propose a summarization method, which
  produces a single text from multiple descriptions. The resultant
  summary concisely describes a term from different viewpoints. We
  also show the effectiveness of our method by means of experiments.
\end{abstractquote}
\end{abstract}

\section{Introduction}
\label{sec:introduction}

Term descriptions, which have been carefully organized in hand-crafted
encyclopedias, are valuable linguistic knowledge for human usage and
computational linguistics research. However, due to the limitation of
manual compilation, existing encyclopedias often lack new terms and
new definitions for existing terms.

The World Wide Web (the Web), which contains an enormous volume of
up-to-date information, is a promising source to obtain new term
descriptions.  It has become fairly common to consult the Web for
descriptions of a specific term. However, the use of existing search
engines is associated with the following problems:
\renewcommand{\theenumi}{\alph{enumi}}
\begin{enumerate}
  \def\labelenumi{(\theenumi)}
\item search engines often retrieve extraneous pages not describing a
  submitted term,
\item even if desired pages are retrieved, a user has to identify page
  fragments describing the term,
\item word senses are not distinguished for polysemous terms, such as
  ``hub (device and center)'',
\item descriptions in multiple pages are independent and do not
  comprise a condensed and coherent text as in existing encyclopedias.
\end{enumerate}

The authors of this paper have been resolving these problems
progressively. For problems (a) and (b), Fujii and
Ishikawa~\shortcite{fujii:acl-2000} proposed an automatic method to
extract term descriptions from the Web.  For problem (c), Fujii and
Ishikawa~\shortcite{fujii:acl-2001} improved the previous method, so
that the multiple descriptions extracted for a single term are
categorized into domains and consequently word senses are
distinguished.

\begin{figure*}[htbp]
  \begin{center}
    \leavevmode
    \includegraphics[height=4.3in]{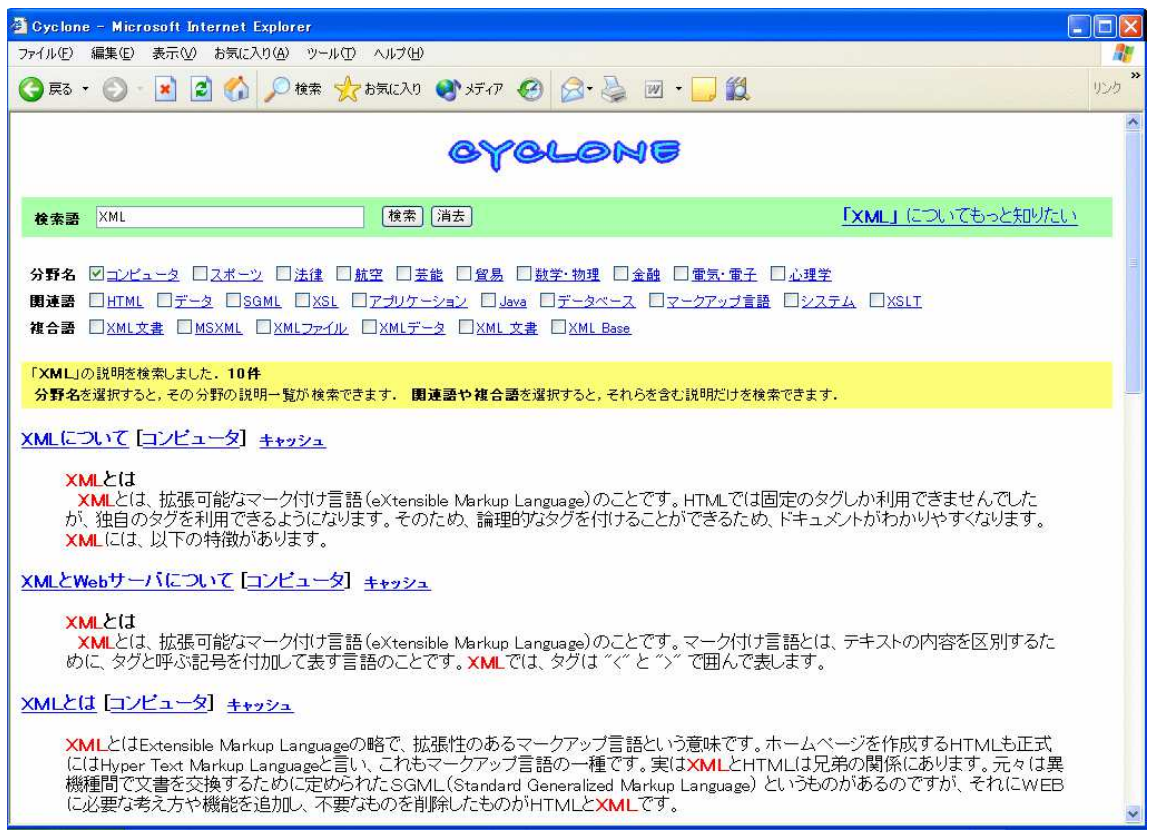}
    \caption{Example descriptions for ``XML''.}
    \label{fig:xml}
  \end{center}
\end{figure*}

Using these methods, we have compiled an encyclopedic corpus for
approximately 600,000 Japanese terms. We have also built a Web site
called ``{\sc Cyclone}''\footnote{http://cyclone.slis.tsukuba.ac.jp/}
to utilize this corpus, in which one or more paragraph-style
descriptions extracted from different pages can be retrieved in
response to a user input. In Figure~\ref{fig:xml}, three paragraphs
describing ``XML'' are presented with the titles of their source pages.

However, the above-mentioned problem (d) remains unresolved and this
is exactly what we intend to address in this paper.

In hand-crafted encyclopedias, a single term is described concisely
from different ``viewpoints'', such as the definition,
exemplification, and purpose.  In contrast, if the first paragraph in
Figure~\ref{fig:xml} is not described from a sufficient number of
viewpoints for XML, a user has to read remaining paragraphs.  However,
this is inefficient, because the descriptions are extracted from
independent pages and usually include redundant contents.

To resolve this problem, we propose a summarization method that
produces a concise and condensed term description from multiple
paragraphs.  As a result, a user can obtain sufficient information
about a term with a minimal cost.  Additionally, by reducing the size
of descriptions, {\sc Cyclone} can be used with mobile devices, such
as PDAs.

However, while {\sc Cyclone} includes various types of terms, such as
technical terms, events, and animals,  the required set of viewpoints
can vary depending the type of target terms. For example, the
definition and exemplification are necessary for technical terms, but
the family and habitat are necessary for animals. In this paper, we
target Japanese technical terms in the computer domain.

Section~\ref{sec:cyclone} outlines {\sc Cyclone}.
Sections~\ref{sec:method}  and \ref{sec:eval} explain our
summarization method and its evaluation, respectively.  In
Section~\ref{sec:discussion}, we discuss related work and the
scalability of our method.

\section{Overview of {\sc Cyclone}}
\label{sec:cyclone}

\begin{figure}[htbp]
  \begin{center}
    \leavevmode
    \includegraphics[height=2.7in]{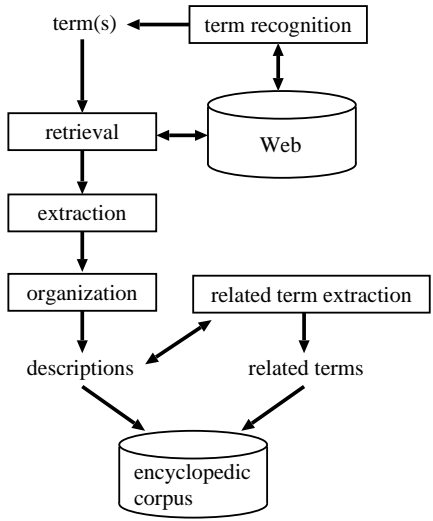}
    \caption{Overall design of {\sc Cyclone}.}
    \label{fig:cyclone}
  \end{center}
\end{figure}

Figure~\ref{fig:cyclone} depicts the overall design of {\sc Cyclone},
which produces an encyclopedic corpus by means of five modules: ``term
recognition'', ``extraction'', ``retrieval'', ``organization'', and
``related term extraction''. While {\sc Cyclone} produces a corpus
off-line, users search the resultant corpus for specific descriptions
on-line.

It should be noted that the summarization method proposed in this
paper is not included in Figure~\ref{fig:cyclone} and that the concept
of viewpoint has not been used in the modules in
Figure~\ref{fig:cyclone}.

In the off-line process, the input terms can be either submitted
manually or collected by the term recognition module automatically.
The term recognition module periodically searches the Web for morpheme
sequences not included in the corpus, which are used as input terms.

The retrieval module exhaustively searches the Web for pages including
an input term, as performed in existing Web search engines.

The extraction module analyzes the layout (i.e., the structure of HTML
tags) of each retrieved page and identifies the paragraphs that
potentially describe the target term. While promising descriptions can
be extracted from pages resembling on-line dictionaries, descriptions
can also be extracted from general pages.

The organization module classifies the multiple paragraphs for a
single term into predefined domains (e.g., computers, medicine, and
sports) and sorts them according to the score. The score is computed
by the reliability determined by hyper-links as in
Google\footnote{http://www.google.com/} and the linguistic validity
determined by a language model produced from an existing
machine-readable encyclopedia.  Thus, different word senses, which are
often associated with different domains, can be distinguished and
high-quality descriptions can be selected for each domain.

Finally, the related term extraction module searches top-ranked
descriptions for terms strongly related to the target term (e.g.,
``cable'' and ``LAN'' for ``hub''). Existing encyclopedias often
provide related terms for each headword, which are effective to
understand the headword.  In {\sc Cyclone}, related terms can also be
used as feedback terms to narrow down the user focus. However, this
module is beyond the scope of this paper.

\section{Summarization Method}
\label{sec:method}

\subsection{Overview}
\label{subsec:overview}

Given a set of paragraph-style descriptions for a single term in a
specific domain (e.g., descriptions for ``hub'' in the computer
domain), our summarization method produces a concise text describing
the term from different viewpoints.

These descriptions are obtained by the organization module in
Figure~\ref{fig:cyclone}.  Thus, the related term extraction module is
independent of our summarization method.

Our method is multi-document summarization (MDS)
\cite{mani:book-2001-chap7}. Because a set of input documents (in our
case, the paragraphs for a single term) were written by different
authors and/or different time, the redundancy and divergence of the
topics in the input are greater than that for single document
summarization. Thus, the recognition of similarity and difference
among multiple contents is crucial. The following two questions have
to be answered:
\begin{itemize}
\item by which language unit (e.g., words, phrases, or sentences)
  should two contents be compared?
\item by which criterion should two contents be regarded as
  ``similar'' or ``different''?
\end{itemize}
The answers for these questions can be different depending on the
application and the type of input documents.

Our purpose is to include as many viewpoints as possible in a concise
description. Thus, we compare two contents on a viewpoint-by-viewpoint
basis. In addition, if two contents are associated with the same
viewpoint, we determine that those contents are similar and that they
should not be repeated in the summary.

Our viewpoint-based summarization (VBS) method consists of the
following four steps: \renewcommand{\theenumi}{\arabic{enumi}}
\begin{enumerate}
\item identification, which recognizes the language unit associated
  with a viewpoint,
\item classification, which merges the identified units associated
  with the same viewpoint into a single group,
\item selection, which determines one or more representative units for
  each group,
\item presentation, which produces a summary in a specific format.
\end{enumerate}
The model is similar to those in existing MDS methods. However, the
implementation of each step varies depending on the application.  We
elaborate on the four steps in
Sections~\ref{subsec:identification}-\ref{subsec:presentation},
respectively.

\subsection{Identification}
\label{subsec:identification}

The identification module recognizes the language units, each of which
describes a target term from a specific viewpoint. However, a compound
or complex sentence is often associated with multiple viewpoints. The
following example is an English translation of a Japanese compound
sentence in a Web page.
\begin{quote}
  XML is an abbreviation for eXtensible Markup Language, and is a
  markup language.
\end{quote}
The first and second clauses describe XML from the abbreviation and
definition viewpoints, respectively. It should be noted that because
``XML'' and ``eXtensible Markup Language'' are spelled out by the
Roman alphabet in the original sentence, the first clause does not
provide Japanese readers with the definition of XML.

To extract the language units on a viewpoint-by-viewpoint basis, we
segment Japanese sentences into simple sentences. However, sentence
segmentation remains a difficult problem and the accuracy is not
100\%.  First, we analyze the syntactic dependency structure of an
input sentence by CaboCha\footnote{\scriptsize
  http://cl.aist-nara.ac.jp/\~{}taku-ku/software/cabocha/}. Second, we
use hand-crafted rules to extract simple sentences using the
dependency structure.

The simple sentences excepting the first clause often lack the
subject. To resolve this problem, zero pronoun detection and anaphora
resolution can be used. However, due to the rudimentary nature of
existing methods, we use hand-crafted rules to complement simple
sentences with the subject.

As a result, we can obtain the following two simple sentences from the
above-mentioned input sentence, in which the complement subject is in
parentheses.
\begin{quote}
  \begin{itemize}
  \item XML is an abbreviation for eXtensible Markup Language.
  \item (XML) is a markup language.
  \end{itemize}
\end{quote}

\subsection{Classification}
\label{subsec:classification}

The classification module merges the simple sentences related to the
same viewpoint into a single group. An existing encyclopedia for
technical terms uses approximately 30 obligatory and optional
viewpoints. We selected the following 12 viewpoints for which typical
expressions can be coded manually:
\begin{quote}
  definition, abbreviation, exemplification, purpose, synonym,
  reference, product, advantage, drawback, history, component,
  function.
\end{quote}
We manually produced 36 linguistic patterns used to describe terms
from a specific viewpoint. These patterns are regular expressions, in
which specific morphemes are generalized into parts-of-speech or the
special symbol representing the target term.

We use a two-stage classification method. First, the simple sentences
that match with a pattern are classified into the associated viewpoint
group. A simple sentence that matches with patterns for multiple
viewpoints is classified into every possible group.

However, the pattern-based method fails to classify the sentences that
do not match with any predefined patterns. Thus, second we classify
the remaining sentences into the group in which the most similar
sentence has already been classified. In practice, we compute the
similarity between an unclassified sentence and each of the classified
sentences. The similarity between two sentences is determined by the
Dice coefficient, i.e., the ratio of content words commonly included
in those sentences.  The sentences unclassified through the above
method are classified into the ``miscellaneous'' group.

In summary, our two-stage method uses predefined linguistic patterns
and statistics of words.

The following examples are English translations of Japanese sentences
extracted in the identification module. These sentences can be
classified into a specific group on the ground of the underlined
expressions, excepting sentence (e). However, in the second stage,
sentence (e) can be classified into the history group, because
sentence (e) is most similar to sentence (c).

\renewcommand{\theenumi}{\alph{enumi}}
\begin{enumerate}
  \def\labelenumi{(\theenumi)}
  
\item $\underline{\mbox{XML is}}$ an extensible markup language.

  $\rightarrow$ definition

\item $\underline{\mbox{an abbreviation for}}$ eXtensible Markup Language

  $\rightarrow$ abbreviation

\item was advised as a standard by W3C $\underline{\mbox{in 1998}}$

  $\rightarrow$ history

\item XML is $\underline{\mbox{an abbreviation for}}$ Extensible Markup
  Language

  $\rightarrow$ abbreviation

\item the standard of XML was advised by W3C

  $\rightarrow$ ???
  $\rightarrow$ history

\end{enumerate}

\subsection{Selection}
\label{subsec:selection}

The selection module determines one or more representative sentences
for each viewpoint group. The number of sentences selected from each
group can vary depending on the desired size of the resultant summary.

We consider the following factors to compute the score for each
sentence and select sentences with greater scores in each group.
\begin{itemize}
\item the number of common words included (W)
  
  The representative sentences should contain many words that are
  common in the group. We collect the frequencies of words for each
  group, and sentences including frequent words are preferred.

\item the rank in {\sc Cyclone} (R)
  
  As depicted in Figure~\ref{fig:cyclone}, {\sc Cyclone} sorts the
  retrieved paragraphs according to the plausibility as the
  description. Sentences in highly-ranked paragraphs are preferred.

\item the number of characters included (C)
  
  To minimize the size of a summary, short sentences are preferred.

\end{itemize}

Because these factors are different in terms of the dimension, range,
and polarity, we normalize each factor in [0,1] and compute the final
score as a weighed average of the three factors. The weight of each
factor was determined by a preliminary study. In brief, the relative
importance among the three factors is W$>$R$>$C.

However, because the miscellaneous group includes various viewpoints,
we use a different method from that for the regular groups.
First, we select representative sentences from the regular groups.
Second, from the miscellaneous group, we select the sentence that is
most dissimilar to the sentences already selected as representatives.
We use the Dice-based similarity used in
Section~\ref{subsec:classification} to measure the dissimilarity
between two sentences.  If we select more than one sentence from the
miscellaneous group, the second process is repeated recursively.

\subsection{Presentation}
\label{subsec:presentation}

The presentation module lists the selected sentences without any
post-editing. Ideally, natural language generation is required to
produce a coherent text by, for example, complementing conjunctions
and generating anaphoric expressions.  However, a simple list of
sentences is also useful to obtain knowledge about a target term.

\begin{figure*}[htbp]
  \begin{center}
    \leavevmode
    \includegraphics[height=4.3in]{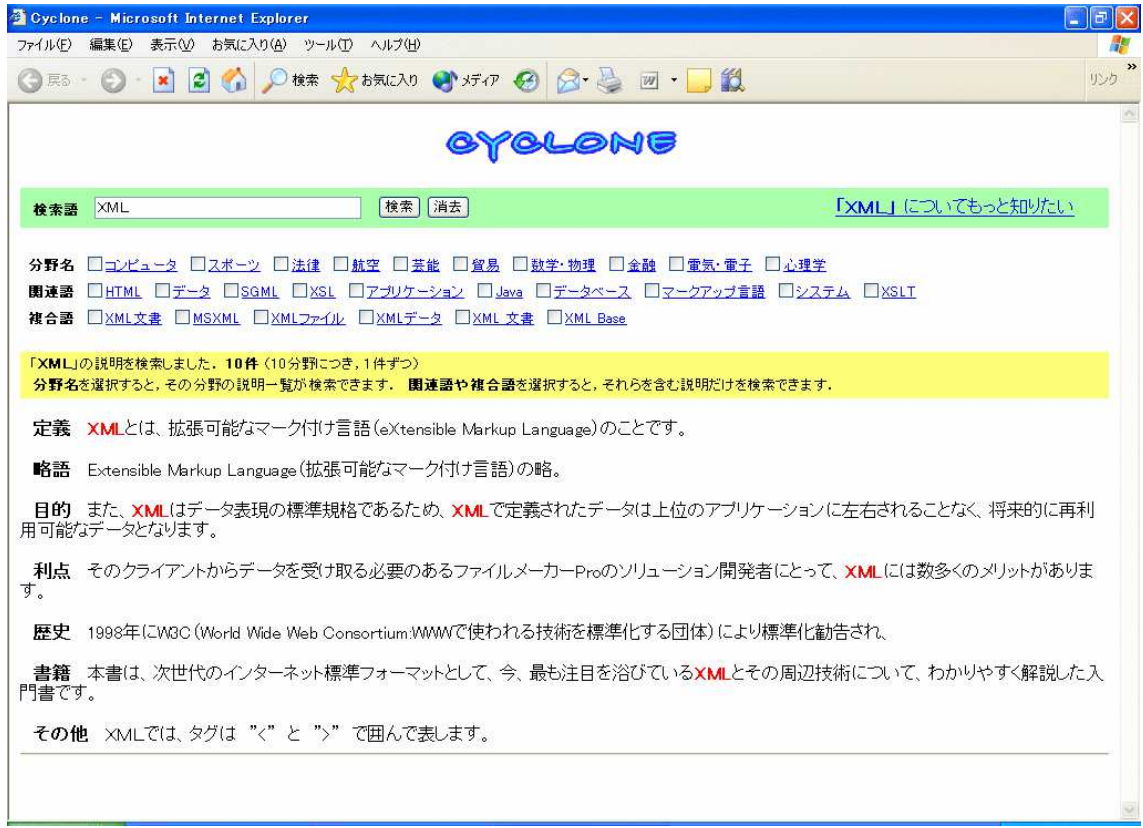}
    \caption{Example summary for ``XML''.}
    \label{fig:xml2}
  \end{center}
\end{figure*}

Figure~\ref{fig:xml2} depicts an example summary produced from the top
50 paragraphs for the term ``XML''. In this figure, six viewpoint
groups and the miscellaneous group were formed and only one sentence
was selected from each group. The order of sentences presented was
determined by the score computed in the selection module.

While the source paragraphs consist of 11,224 characters, the summary
consists of 397 characters, which is almost the same length as an
abstract for a technical paper.

The following is an English translation of the sentences in
Figure~\ref{fig:xml2}. Here, the words spelled out by the Roman
alphabet in the original sentences are in italics.

\begin{itemize}
\item {\bf definition}: {\it XML\/} is an extensible markup language ({\it
    eXtensible Markup Language\/}).
\item {\bf abbreviation}: an abbreviation for {\it Extensible Markup
    Language\/} (an extensible markup language).
\item {\bf purpose}: Because {\it XML\/} is a standard specification
  for data representation, the data defined by {\it XML\/} can be
  reusable, irrespective of the upper application.
\item {\bf advantage}: {\it XML\/} is advantageous to developers of
  the file maker {\it Pro\/}, which needs to receive data from the
  client.
\item {\bf history}: was advised as a standard by {\it W3C\/} ({\it
    World Wide Web Consortium\/}: a group standardizing {\it WWW\/}
  technologies) in 1998,
\item {\bf reference}: This book is an introduction for {\it XML\/},
  which has recently been paid much attention as the next generation
  Internet standard format, and related technologies.
\item {\bf miscellaneous}: In {\it XML\/}, the tags are enclosed in
  ``$<$'' and ``$>$''.
\end{itemize}
Each viewpoint label or sentence is hyper-linked to the associated
group or the source paragraph, respectively, so that a user can easily
obtain more information on a specific viewpoint. For example, by the
reference sentence, a catalogue page of the book in question can be
retrieved.

Although the resultant summary describes XML from multiple viewpoints,
there is a room for improvement. For example, the sentences classified
into the definition and abbreviation viewpoints include almost the
same content.

\section{Evaluation}
\label{sec:eval}

\subsection{Methodology}
\label{subsec:eval_method}

Existing methods for evaluating summarization techniques can be
classified into intrinsic and extrinsic approaches.

In the intrinsic approach, the content of a summary is evaluated with
respect to the quality of a text (e.g., coherence) and the
informativeness (i.e., the extent to which important contents are in
the summary).  In the extrinsic approach, the evaluation measure is
the extent to which a summary improves the efficiency of a specific
task (e.g., relevance judgment in text retrieval).

In DUC\footnote{http://duc.nist.gov/} and
NTCIR\footnote{http://research.nii.ac.jp/ntcir/index-en.html}, both
approaches have been used to evaluate summarization methods targeting
newspaper articles. However, because there was no public test
collections targeting term descriptions in Web pages, we produced our
test collection. As the first step of our summarization research, we
addressed only the intrinsic evaluation.

In this paper, we focused on including as many viewpoints (i.e.,
contents) as possible in a summary, but did not address the text
coherence. Thus, we used the informativeness of a summary as the
evaluation criterion.  We used the following two measures, which are
in the trade-off relation.
\begin{itemize}
\item compression ratio

  $\frac{\textstyle \mbox{\#characters in summary}}{\textstyle
  \mbox{\#characters in {\sc Cyclone} result}}$
\item coverage

  $\frac{\textstyle \mbox{\#viewpoints in summary}}{\textstyle
  \mbox{\#viewpoints in {\sc Cyclone} result}}$
\end{itemize}
Here, ``\#viewpoints'' denotes the number of viewpoint {\it types\/}.
Even if a summary contains multiple sentences related to the same
viewpoint, the numerator is increased by 1.

We used 15 Japanese term in an existing computer dictionary as test
inputs. English translations of the test inputs are as follows:
\begin{quote}
  10BASE-T, ASCII, SQL, XML, accumulator, assembler, binary number,
  crossing cable, data warehouse, macro virus,  main memory unit,
  parallel processing, resolution, search time, thesaurus.
\end{quote}

To calculate the coverage, the simple sentences in the {\sc Cyclone}
results have to be associated with viewpoints.  To reduce the
subjectivity in the evaluation, for each of the 15 terms, we asked two
college students (excluding the authors of this paper) to annotate
each simple sentence in the top 50 paragraphs with one or more
viewpoints. The two annotators performed the annotation task
independently. The denominators of the compression ratio and coverage
were calculated by the top 50 paragraphs.

During a preliminary study, the authors and annotators defined 28
viewpoints, including the 12 viewpoints targeted in our method.  We
also defined the following three categories, which were not considered
as a viewpoint:
\begin{itemize}
\item non-description, which were also used to annotate non-sentence
  fragments caused by errors in the identification module,
\item description for a word sense independent of the computer domain
  (e.g., ``hub'' as a center, instead of a network device),
\item miscellaneous.
\end{itemize}

It may be argued that an existing hand-crafted encyclopedia can be
used as the standard summary. However, paragraphs in {\sc Cyclone}
often contain viewpoints not described in existing
encyclopedias. Thus, we did not use existing encyclopedias in our
experiments.

\subsection{Results}
\label{subsec:results}

Table~\ref{tab:results} shows the compression ratio and coverage for
different methods, in which ``\#Reps'' and ``\#Chars'' denote the
number of representative sentences selected from each viewpoint group
and the number of characters in a summary, respectively. We always
selected five sentences from the miscellaneous group.  The third
column denotes the compression ratio.

\begin{table*}[htbp]
  \begin{center}
    \caption{Results of summarization experiments.}
    \medskip
    \leavevmode
    \small
    \begin{tabular}{ccccccccccc} \hline\hline
      & &
      & \multicolumn{4}{c}{Coverage by annotator A (\%)}
      & \multicolumn{4}{c}{Coverage by annotator B (\%)} \\
      \cline{4-7} \cline{8-11}
      & & Compression
      & \multicolumn{2}{c}{12 Viewpoints}
      & \multicolumn{2}{c}{28 Viewpoints}
      & \multicolumn{2}{c}{12 Viewpoints}
      & \multicolumn{2}{c}{28 Viewpoints} \\
      \cline{4-5} \cline{6-7} \cline{8-9} \cline{10-11}
      \#Reps & \#Chars &
      {\hfill\centering ratio (\%)\hfill} &
      {\hfill\centering VBS \hfill} &
      {\hfill\centering Lead \hfill} &
      {\hfill\centering VBS \hfill} &
      {\hfill\centering Lead \hfill} &
      {\hfill\centering VBS \hfill} &
      {\hfill\centering Lead \hfill} &
      {\hfill\centering VBS \hfill} &
      {\hfill\centering Lead \hfill} \\
      \hline
      1 & 616 & 5.97 & 56.62 & 52.84 & 49.49 & 44.84 & 50.00 & 53.61 &
      49.49 & 47.56 \\
      2 & 998 & 9.61 & 73.43 & 57.23 & 59.26 & 53.70 & 64.50 & 62.96
      & 60.75 & 57.37 \\
      3 & 1309 & 12.61 & 76.04 & 59.29 & 63.13 & 56.44 & 67.83 & 64.81
      & 65.22 & 60.84 \\
      \hline
    \end{tabular}
    \label{tab:results}
  \end{center}
\end{table*}

The remaining columns denote the coverage on a annotator-by-annotator
basis. The columns ``12 Viewpoints'' and ``28 Viewpoints'' denote the
case in which we focused only on the 12 viewpoints targeted in our
method and the case in which all the 28 viewpoints were considered,
respectively.

The columns ``VBS'' and ``Lead'' denote the coverage obtained with our
viewpoint-based summarization method and the lead method. The lead
method, which has often been used as a baseline method in past
literature, systematically extracted the top $N$ characters from the
{\sc Cyclone} result. Here, $N$ is the same number in the second
column.

In other words, the compression ratio of the VBS and lead methods was
standardized, and we compared the coverage of both methods.  The
compression ratio and coverage were averaged over the 15 test terms.

Suggestions which can be derived from Table~\ref{tab:results} are as
follows.

First, in the case of ``\#Reps=1'', the average size of a summary was
616 characters, which is marginally longer than an abstract for a
technical paper. In the case of ``\#Reps=3'', the average summary size
was 1309 characters, which is almost the maximum size for a single
description in hand-crafted encyclopedias. A summary obtained with
four sentences in each group is perhaps too long as term descriptions.

Second, the compression ratio was roughly 10\%, which is fairly good
performance. It may be argued that the compression ratio is
exaggerated. That is, although paragraphs ranked higher than 50 can
potentially provide the sufficient viewpoints, the top 50 paragraphs
were always used to calculate the dominator of the compression ratio.

We found that the top 38 paragraphs, on average, contained all
viewpoint types in the top 50 paragraphs. Thus, the remaining 12
paragraphs did not provide additional information.  However, it is
difficult for a user to determine when to stop reading a retrieval
result. In existing evaluation workshops, such as NTCIR, the
compression ratio is also calculated using the total size of the input
documents.

Third, the VBS method outperformed the lead method in terms of the
coverage, excepting the case of ``\#Reps=1'' focusing on the 12
viewpoints by annotator B.  However, in general the VBS method
produced more informative summaries than the lead method, irrespective
of the compression ratio and the annotator.

It should be noted that although the VBS method targets 12 viewpoints,
the sentences selected from the miscellaneous group can be related to
the remaining 16 viewpoints. Thus, even if we focus on the 28
viewpoints, the coverage of the VBS method can potentially increase.

It should also be noted that all viewpoints are not equally
important. For example, in an existing
encyclopedia~\cite{nagao:iwanami-1990} the definition,
exemplification, and synonym are regarded as the obligatory
viewpoints, and the remaining viewpoints are optional.

We investigated the coverage for the three obligatory viewpoints. We
found that while the coverage for the definition and exemplification
ranged from 60\% to 90\%, the coverage for the synonym was 50\% or
less.

A low coverage for the synonym is partially due to the fact that
synonyms are often described with parentheses.  However, because
parentheses are used for various purposes, it is difficult to identify
only synonyms expressed with parentheses. This problem needs to be
further explored.

\section{Discussion}
\label{sec:discussion}

The goal of our research is to automatically compile a high-quality
large encyclopedic corpus using the Web. Hand-crafted encyclopedias
lack new terms and new definitions for existing terms, and thus the
quantity problem is crucial. The Web contains unreliable and
unorganized information and thus the quality problem is crucial. We
intend to alleviate both problems.  To the best of our knowledge, no
attempt has been made to intend similar purposes.

Our research is related to question answering (QA). For example, in
TREC QA track, definition questions are intended to provide a user
with the definition of a target item or
person~\cite{voorhees:hlt-naacl-2003}.  However, while the expected
answer for a TREC question is short definition sentences as in a
dictionary, we intend to produce an encyclopedic text describing a
target term from multiple viewpoints.

The summarization method proposed in this paper is related to
multi-document summarization (MDS)
\cite{mani:book-2001-chap7,radev:cl-1998,schiffman:acl-2001}. The
novelty of our research is that we applied MDS to producing a
condensed term description from unorganized Web pages, while existing
MDS methods used newspaper articles to produce an outline of an event
and a biography of a specific person. We also proposed the concept of
viewpoint for MDS purposes.

While we targeted Japanese technical terms in the computer domain, our
method can also be applied to other types of terms in different
languages, without modifying the model. However, a set of viewpoints
and patterns typically used to describe each viewpoint need to be
modified or replaced depending the application. Given annotated data,
such as those used in our experiments, machine learning methods can
potentially be used to produce a set of viewpoints and patterns for a
specific application.

\section{Conclusion}
\label{sec:conclusion}

To compile encyclopedic term descriptions from the Web, we introduced
a summarization method to our previous work. Future work includes
generating a coherent text instead of a simple list of sentences and
performing extensive experiments including an extrinsic evaluation
method.

\bibliographystyle{acl}

\end{document}